\pdfoutput=1

\documentclass[11pt]{article}

\usepackage{ACL}

\usepackage{times}
\usepackage{latexsym}

\usepackage[T1]{fontenc}

\usepackage[utf8]{inputenc}

\usepackage{microtype}

\usepackage{inconsolata}

\usepackage{amsmath}
\usepackage{amsfonts}
\usepackage{graphicx}
\usepackage{subcaption}
\usepackage{tikz}
\usepackage{multirow}
\usepackage{float}
\usepackage{booktabs}
\usepackage{tcolorbox}
\usepackage{float}
\usepackage{multicol}
\usepackage{algorithm}
\usepackage{algpseudocode}
\usepackage{soul}
\usepackage{arydshln}
\usepackage{bbm}
\usetikzlibrary{bayesnet}

\title{Guiding Natural Language Inference through \\ Lexical Inference Types}

\author{Yingji Zhang$^{1\dagger}$,~ Danilo S. Carvalho$^{1}$,~ Ian Pratt-Hartmann$^{1}$,~ Andr\'{e} Freitas$^{1,2}$ \\
  Department of Computer Science, University of Manchester, United Kingdom$^{1}$ \\
  Idiap Research Institute, Switzerland$^{2}$ \\
  \texttt{\{firstname.lastname\}@[postgrad.]$^{\dagger}$manchester.ac.uk}}

\begin{document}
\maketitle
\begin{abstract}
Explanatory natural language inference (NLI) aims to provide a mechanism to produce step-wise explanations by linking premises to conclusions. While previous research efforts have concentrated on generating unconstrained entailment trees, they lack the design of mechanisms for inference control, where the linguistic and symbolic properties of explanations can be guided and better localised within the latent space of the neural NLI model. To improve symbolic control and localisation, in this paper, we start by systematically characterising the symbolic mechanisms behind explanatory sentences and defining a set of explanatory inference types. Next, we formalise the encoder-decoder NLI model as a conditional latent variable model where the dynamics of latent variables can be guided and manipulated via inference types. Empirical results show that the proposed approach can improve inference and retrieval and deliver symbolic control.
\end{abstract}
\section{Introduction}

\begin{figure}[t]
    \centering
    \includegraphics[scale=0.26]{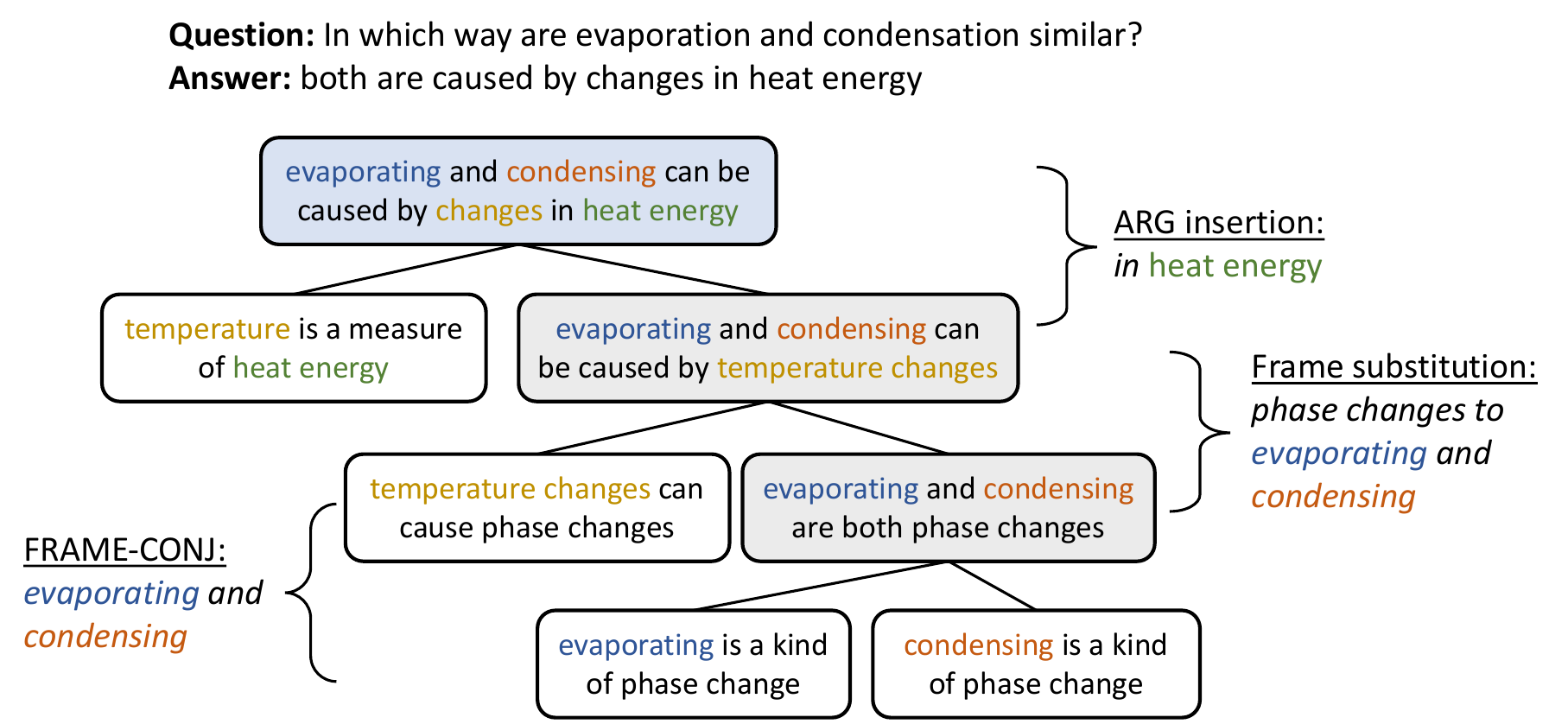}
    \caption{Example of entailment tree. Three steps (1, 2, 3) are present from bottom to top. Each step shows a clear symbolic inference behaviour.}
    \label{fig:tree}
\end{figure}

Explanation-based Natural Language Inference (NLI) aims to provide a mechanism to produce explanatory (abductive) inference chains which ground claims to their supporting premises \cite{https://doi.org/10.48550/arxiv.2010.00389}. This inference process allows a systematic way to support the construction of a step-wise explanation hierarchy, via a particular type of textual entailment which encodes claim-premise relationships at different levels of abstraction. This type of inference is particularly relevant in scientific domains, where reasoning needs to be grounded on sets of known facts and plausible and verifiable abstraction steps. One common way of organising these explanatory chains is with the support of entailment trees (Figure \ref{fig:tree}), a hierarchical structure linking multiple layers of multiple premises to claims. e.g., in the EntailmentBank \cite{dalvi2021explaining}.

Although Transformer-based language models \cite{raffel2020exploring} have been a fundamental component of explanation-based NLI models due to their transferability across different NLI tasks \cite{https://doi.org/10.48550/arxiv.2010.00389}, the controllability and step-wise interpretability of its internal reasoning have not been fully addressed \cite{abzianidze2023formal}. Additionally, one compounding complexity factor to the problem is the fact that natural language explanatory inference is not fully formalisable within a logical setting, being at the interface between formal and material (content-based) inference \cite{valentino2021natural}.

Delivering a controlled explanation-based NLI from a language model thus requires first the characterisation of the formal/symbolic components of explanatory reasoning, from which inference control mechanisms can be derived. For example, within each premise-claim step in Figure \ref{fig:tree}, there are observable symbolic patterns of inference such as substitution (step 2), conjunction (step 1), and specification (step 3). 
Those lexical inference patterns (named here \textit{inference types}) reflect a step-wise/localised semantic change that can guide a language model to perform explainable inference in a well-defined, controllable manner, while keeping content-wise flexibility. Specifically, we claim that a representation model aligned to inference types would allow for improved control and further latent space interpretability. In this work, we focus on the challenge of delivering symbolic control on single-step explanatory NLI (syllogistic-deductive NLI), by targeting the following contributions:

\textbf{-} \textit{A systematic characterisation of explanatory inference types.} To deliver a more controllable and interpretable explanation-based NLI at each entailment step, we start by systematically characterising a set of symbolic patterns and defining associated lexical inference types. The patterns are defined and grounded over a predicate argument-structure representation (Abstract Meaning Representation - AMR symbolic graph) \cite{banarescu2013abstract}.

\textbf{-} \textit{New NLI mechanisms for integrating inference types within a language model.} Next, to bridge the symbolic and distributional representations, we formalise the encoder-decoder NLI model (T5) as a conditional latent variable model where the inference dynamics over latent variables can be guided and manipulated by our defined inference types. As a complementary, we provide an initialisation of the NLI model with sentence space (named T5bottleneck) to induce the inference information in a low-dimensional latent sentence space.

The empirical analysis indicates that eliciting the inference patterns can assist the training, inference, and in-context learning of the explanatory NLI model. T5bottleneck can improve the explanatory inference retrieval \cite{valentino2021hybrid} compared with other latent variable baselines. Moreover, the learned representation can separate different inference types and their symbolic transformation behaviour, indicating the inference rule is encoded in the latent space of the NLI model, implying a quasi-symbolic inference behaviour to the model, where symbolic formal patterns are used in coordination with content-based (distributional) inference. 
\section{Related Work} \label{sec:related}
In this section, we review the related work around three topics: \textit{neuro-symbolic NLI and concepts}, \textit{latent variable model}, \textit{condition(labels)}, to further illustrate our idea and motivation.
\paragraph{Neural-symbolic NLI and concepts.} A long-term goal in NLP is to integrate the representation power of neural networks with the interpretability of symbolic or logical systems to build robust NLI models. Current methods often inject symbolic behavior through explicit symbolic representations, including graph \cite{khashabi2018question,khot2017answering,jansen2017framing,kalouli-etal-2020-hy,thayaparan2021explainable}, linear programming \cite{valentino2022case,thayaparan2024differentiable}, adopting iterative methods, using sparse encoding mechanisms \cite{valentino2020explainable,lin2020differentiable}, synthetic natural language expression \cite{clark2020transformers,yanaka-etal-2021-sygns,fu2024exploring}, symbolic-refined LLMs \cite{olausson-etal-2023-linc,quan2024verification}, etc. 

Recent studies have shown that neural networks can encode sparse neural-symbolic concepts without explicit symbolic injection, across areas like image embedding \cite{ren2022towards,deng2021discovering,li2023does}, word embedding \cite{ethayarajh2018towards,allen2019vec,ri2023contrastive}, contextual embedding \cite{gurnee2023finding,nanda-etal-2023-emergent,park2023linear,li2024inference}, and LLM interpretation \cite{templeton2024scaling}. However, whether transformer-based NLI models can exhibit symbolic behavior remains underexplored, and this is the focus of our work.

\paragraph{Latent variable models in NLP.} Latent variable models, such as VAE \cite{kingma2013auto}, have shown the capability of symbolic representation, control, and interpretation over the latent sentence space where the dimensions are orthogonality and features are casually separated, which are widely deployed in the NLP domain, such as disentangled representation learning \cite{zhang2023learning,zhang-etal-2024-graph}, style-transfer \cite{liu-etal-2023-composable,gu-etal-2023-controllable}, etc. Comparatively, latent variable models are less utilised within NLI settings where fine-grained symbolic control is required \cite{zhang-etal-2024-improving}. Thus, we promote this general concept by eliciting the latent variable model potential inherent in contemporary gradient-based NLI models.
\paragraph{Latent symbolic control via label.} One approach to achieving symbolic control in latent variable models involves conditioning on well-defined domain features, using architectures like CVAEs \cite{carvalho2023learning} and diffusion models \cite{dhariwal2021diffusion,ho2022classifier}. These methods allow for localized generation by injecting labels into the latent space, a technique widely used in the image domain. Similarly, in the mathematical inference domain \cite{meadows2023symbolic}, label information is needed to guide the generation of conclusions from premise expressions. Inspired by those studies, we propose to deliver symbolic NLI control through label-based inference behaviour. In the next section, we are going to introduce the labels (inference types).

\begin{table*}[ht!]
    \small
    \centering
    \resizebox{15.6cm}{!}{
    \begin{tabular}{p{3cm}p{2.8cm}p{0.5cm}p{8.2cm}}
    \toprule
        Original type & AMR type & Prop. & Example entailment relation \\ \hline
        \multirow{9}{*}{Substitution} & \multirow{3}{*}{\shortstack{ARG substitution \\ (ARG-SUB)}} & \multirow{3}{*}{19\%} & P1: \textcolor{blue}{a scar on the knee} is a kind of \textcolor{red}{scar} \\
        &&& P2: a \textcolor{red}{scar} is an acquired characteristic \\
        &&& C: \textcolor{blue}{a scar on the knee} is an acquired characteristic \\
        & \multirow{3}{*}{\shortstack{PRED substitution \\ (PRED-SUB)}} & \multirow{3}{*}{5\%} & P1: food \textcolor{red}{contains} nutrients and energy for living things \\
         &&& P2: to \textcolor{red}{contain} something can mean to \textcolor{blue}{store} something \\
          &&& C: food \textcolor{blue}{stores} nutrients and energy for living things \\
        & \multirow{3}{*}{\shortstack{Frame substitution \\ (FRAME-SUB)}} & \multirow{3}{*}{20\%}  & P1: the \textcolor{blue}{formation of diamonds} requires \textcolor{red}{intense pressure} \\
         &&& P2: the \textcolor{red}{pressure is intense} deep below earth 's crust \\
          &&& C: the \textcolor{blue}{formation of diamonds} occurs deep below the crust of the earth \\ \hline
        \multirow{3}{*}{\shortstack{Inference from Rule}} & \multirow{3}{*}{\shortstack{Conditional frame \\ insertion/substitution \\ (COND-FRAME)}} & \multirow{3}{*}{12\%}  & P1: if \textcolor{blue}{something is renewable} then \textcolor{red}{that something is not a fossil} \\ 
        &&& P2: \textcolor{blue}{fuel wood is a renewable resource} \\
        &&& C: \textcolor{red}{wood is not a fossil fuel} \\ \hline
        \multirow{6}{*}{\shortstack{Further Specification \\ or Conjunction}} & \multirow{3}{*}{\shortstack{ARG insertion \\ (ARG-INS)}} & \multirow{3}{*}{18\%}  & P1: solar energy \textcolor{blue}{comes from the sun} \\ 
        &&& P2: \textcolor{red}{solar energy is a kind of energy} \\
        &&& P3: \textcolor{red}{solar energy is a kind of energy} that \textcolor{blue}{comes from the sun} \\
        & \multirow{3}{*}{\shortstack{Frame conjunction \\ (FRAME-CONJ)}} & \multirow{3}{*}{6\%} & P1: \textcolor{blue}{photosynthesis stores energy} \\
        &&& P2: \textcolor{red}{respiration releases energy} \\
        &&& C: \textcolor{blue}{photosynthesis stores energy} and \textcolor{red}{respiration releases energy} \\ \hline
        \multirow{3}{*}{\shortstack{Infer Class \\ from Properties}} & \multirow{3}{*}{\shortstack{ARG/PRED \\ generalisation \\ (ARG/PRED-GEN)}} & \multirow{3}{*}{1\%}  & P1: \textcolor{blue}{rock} is a hard material \\ 
        &&& P2: \textcolor{red}{granite} is a hard material \\
        &&& C: \textcolor{red}{granite} is a kind of \textcolor{blue}{rock} \\ \hline
        \multirow{3}{*}{\shortstack{Property Inheritance}} & \multirow{3}{*}{\shortstack{ARG substitution \\ (Property Inheritance) \\ (ARG-SUB-PROP)}} & \multirow{3}{*}{0.4\%}  & P1: \textcolor{blue}{blacktop} is made of asphalt concrete \\ 
        &&& P2: asphalt \textcolor{red}{has a smooth surface} \\
        &&& C: a \textcolor{blue}{blacktop} \textcolor{red}{has a smooth surface} \\ \hline

\multirow{9}{*}{Unknown} & \multirow{3}{*}{Example (EXAMPLE)} & \multirow{3}{*}{0.9\%} & a shelter can be used for living in by raccoons \\ 
&&& some raccoons live in hollow logs \\
&&& \textcolor{blue}{an example of} a shelter is a raccon living in a hollow log \\
\specialrule{0pt}{2pt}{2pt}
& \multirow{3}{*}{If ... then ... (IFT)} & \multirow{3}{*}{0.8\%} & an optical telescope requires visible light for human to use \\ 
&&& clouds / dusts block visible light \\
&&& \textcolor{blue}{if} there is clouds or dusts, \textcolor{blue}{then} the optical telescope cannot be used  \\
\specialrule{0pt}{2pt}{2pt}
& \multirow{3}{*}{Others (UNK)} & \multirow{3}{*}{16\%} & spiral is a kind of shape \\ 
&&& galaxies can be classified by shape \\
&&& spiral galaxy is a type of galaxy \\ \toprule
\end{tabular}
    }
\caption{Example of AMR-based inference types. Their abbreviations are described in the parentheses, which is used in the paper. The size of the EntailmentBank is 5134 in our task. The annotations are going to be released.} \label{tab:inference_type_example}
\end{table*}
\section{Defining Lexical Inference Types} \label{sec:bg}
\citet{valentino2021natural} has demonstrated that step-wise explanation-based NLI cannot be directly framed as pure logical reasoning. Explanatory chains, while looking plausible at first inspection, commonly have subtler incompleteness and consistency problems from a logical point of view. Meanwhile, explanatory chains correspond to definable inference patterns and symbolic operations can be localised over the sentence structure. Motivated by this middle ground between logical representations and lexico-semantic inference patterns, 
we introduce granular inference types based on explanatory sentences, using AMR to define the symbolic operations involved in step-wise inference, linking transformations from premises to conclusions \footnote{Please note that AMR is not used as a representation mechanism in the proposed architecture, but only to precisely ground these symbolic operations within a well-defined semantic representation structure.}.
Table \ref{tab:inference_type_example} describes the AMR-grounded inference types and examples from the EntailmenBank corpus. Next, we define each lexico-semantic inference type and the corresponding symbolic forms.

\begin{figure}[ht!]
    \centering
    \includegraphics[scale=0.47]{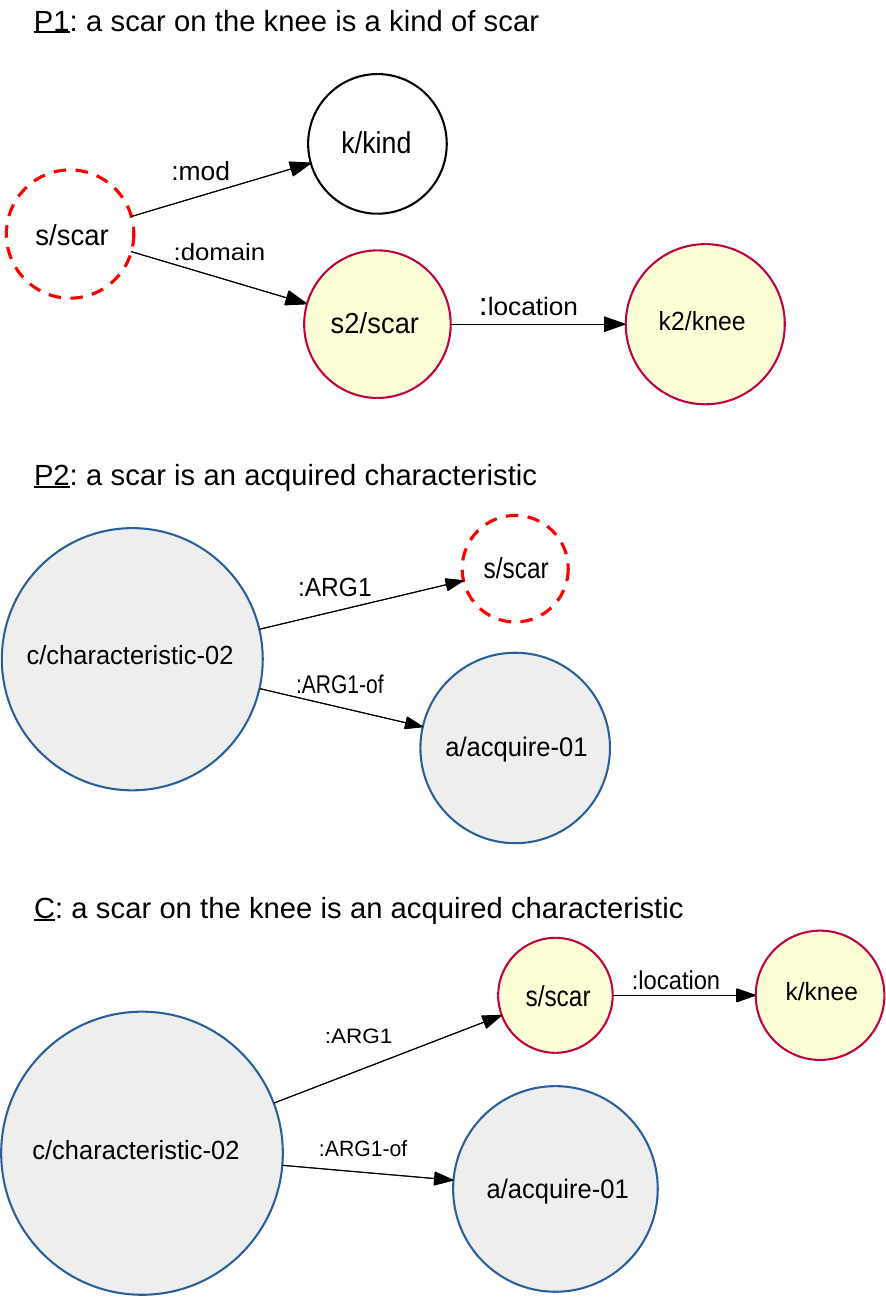}
    \caption{AMR argument substitution: the inference behaviour is defined as subgraph substitution.}
    \label{fig:amr_argsub}
\end{figure}

The substitution category refers to obtaining a conclusion by replacing a predicate/argument term from one premise with a predicate/argument term from the other premise. Possible variations of this category include (1) \textit{argument (ARG) substitution}, (2) \textit{predicate (PRED) substitution}, and (3) \textit{frame (PRED+ARG) substitution}. In this category, one premise is used to connect two terms which are usually connected by \textit{is a kind of, is a source of}, etc. Conceptualising the AMR representation as a graph, this can be symbolically represented as a subgraph substitution operation over the premise graphs, as illustrated in Figure~\ref{fig:amr_argsub}. The \textit{PRED substitution} category works in a similar manner, but replacing a predicate term. The two predicates are usually linked by the following patterns: ``\textit{$v_1$ is a kind of $v_2$}'', ``\textit{to $v_1$ something means to $v_2$ something}'', etc. The \textit{frame (PRED+ARG) substitution} category combines both previous categories by replacing a frame (predicate subgraph) of one of the premises with one from the other premise.

The \textit{further specification} or \textit{conjunction} category allows for obtaining a conclusion by joining both premises. It includes (4) \textit{ARG insertion} and (5) \textit{frame conjunction}. In the case of \textit{ARG insertion}, the conclusion is obtained by connecting an argument from one of the premises to a frame of the other.
As for \textit{frame conjunction/disjunction}, the conclusion is obtained by joining the premises graphs through a conjunction/disjunction node (\textit{and}) or (\textit{or}). The \textit{inference from rule} category from \cite{dalvi2021explaining} comprises a specific case of insertion or substitution categorised as (6) \textit{conditional frame insertion/substitution}, where a frame is inserted or replaced as an argument of a premise following a conditional path in the other premise, as illustrated in Figure~\ref{fig:amr_condframesub}.

The inference type \textit{infer class from properties} has been re-categorised as (7) \textit{ARG or PRED generalisation}, where a new \textit{:domain} relation frame is created if both premise graphs differ by a single predicate/argument term.
(8) \textit{Property inheritance}, on the other hand, is a special case of \textit{ARG substitution}, where one of the premises describes a \textit{is made of} relationship between the entity in the other premise and its replacement. 
Finally, the \textit{other inference types} category includes (9) \textit{example}, (10) \textit{if-then}, and (11) \textit{others}. Those inference types are defined according to the key lexical characteristic of the conclusion, as systematic AMR transformations which could be applied without rephrasing the underlying explanatory sentences could not be determined. More details about this category and annotation procedure are provided in Appendix \ref{sec:annotation}.

\section{Latent Variable NLI Framework} \label{sec:latent_props}

In this section, we first frame the NLI model as a latent variable model and align the notion of inference types as a conditional mechanism to support the learning and control of a quasi-symbolic latent space in section \ref{sec:latent_var_NLI}. Then, we instantiate the mechanism within a supporting neural architecture for learning latent sentence space in section \ref{sec:latent_var_NLI_arc}.

\subsection{Latent variable NLI} \label{sec:latent_var_NLI}
Recent work revealed that transformer-based language models can linearly encode abstract-level semantic concepts (latent variables, denoted by $z$) \cite{park2023linear,li2024inference,wang2024large,jiang2024origins}. Following prior studies, we frame gradient-based neural NLI models as conditional latent variable models that can realise a quasi-symbolic inference dynamics. Assuming premises and conclusions share the same latent space where the explanatory entailment relation is computed in a probabilistic fashion, this allows for the framing of the entailment determination as
the problem of learning a set of conditional probabilities among the latent variables. Figure \ref{fig:formalization} depicts an abstraction of the computational graph of the latent NLI/explanatory entailment framework.

\begin{figure}[ht!]
\centering
\resizebox{7.8cm}{!}{
\begin{minipage}{3cm}
\begin{tikzpicture}
  \node[latent] (z) at (0,2) {$z$};
  \node[latent] (x) at (0,0) {$x$};
  \edge {z} {x};
  \path (z) edge[->] node[left] {$p_\theta(x|z)$} (x);
\end{tikzpicture}
\subcaption{Latent NLI model}
\end{minipage}
\hspace{1cm}
\begin{minipage}{4cm}
\begin{tikzpicture}
  \node[latent] (r) at (2.4,2) {$\pi$};
  \node[latent] (z) at (0,2) {$z$};
  \node[latent] (x) at (1.2,0) {$x$};
  \edge {r} {z};
  \edge {z,r} {x};
  \path (r) edge[->] node[below] {$p_\theta(z|\pi)$} (z);
  \path (z) edge[->] node[below,sloped] {\small $p_\theta(x|z,\pi)$} (x);
\end{tikzpicture}
\subcaption{Conditional NLI model}
\end{minipage}
\hspace*{1pt}
}
\caption{Latent variable NLI framework.}
\label{fig:formalization}
\end{figure}

\paragraph{Latent variables and relations.} We first propose a set of latent variables based on prior studies of \citet{zhang2023learning}, which revealed that explanatory sentence semantics can be decomposed into \textit{semantic role - word content} sets (denoted by role-content) according to Argument Structure Theory (AST) \cite{jackendoff1992semantic}. E.g., the sentence, `animals require oxygen for survival', can be represented as:
$\underbrace{animals}_{ARG0}\oplus \underbrace{require}_{PRED}\oplus\underbrace{oxygen}_{ARG1}\oplus\underbrace{for~survival}_{ARGM-PRP}$
\noindent where $\oplus$ represents the composition operation under a compositional-distributional model  \cite{clark2008compositional}. Each role-content set, such as \textit{ARG0-animals}, is encoded as a convex cone in the latent space. Therefore, we consider each role-content as a latent variable. The latent representation of observed sentence $x$ can be formalised as a set of latent variables:
$x \leftrightarrow z^{(x)} = \{ ({c_1}, {r_1}) , \dots , ({c_{i}},{r_{i}}) , \dots \}$
where $\leftrightarrow$ represent the deterministic mapping between $x$ and $z^{(x)}$ through the embedding layer, $c_i \in C$ and $r_i \in R$ represent the word content and semantic role at position $i$, $C$ and $R$ are the vocabularies of word content and semantic role category, predefined based on training corpus. 

Since an AMR representation is a particular instance of an AST representation, which represents the relation between latent variables, by defining and manipulating the inference patterns over the AMR representation in the context of inference types, we can provide quasi-symbolic interpretation and control to the latent NLI model. In the next section, the targeted NLI task supported by the AMR-grounded inference types is formalised under a Bayesian inference framework.


\paragraph{Latent Bayesian inference.} 

In the context of step-wise, AMR-grounded explanations, 
given a premise-conclusion explanatory sentence pair $<x_{p_0}, x_{p_1}, x_c>$, an inference type $\pi \in \Pi$ can be associated, if exists a transformation $amr(x_{p_0}), amr(x_{p_1}) \rightarrow amr(x_c)$ defined over the set of transformations $\Pi$.
The NLI process can be described as a Bayesian inference:
$
P(x_{c}|x_{p_{0}}, x_{p_{1}}) = P(x_{c}|z^{(x_c)}) P(z^{(x_c)}|x_{p_{0}}, x_{p_{1}})
$
where $P(z^{(x_c)}|x_{p_{0}}, x_{p_{1}})$ approximates the posterior inference via the encoder. Specifically, it first transform $x_{p_{0}}, x_{p_{1}}$ into latent representations $z^{x_{p_{0}}}, z^{x_{p_{1}}}$. Subsequently, inference behaviour $\pi$ is performed over the set of latent variables (e.g., substitution over latent variables set). The latent variables $z^{(x_{c})}$ are retained for generation conclusion via decoder $P(x_{c}|z)$. To validate this inference process, we propose Proposition 1.

\textit{\textbf{Proposition 1:}} \textit{The inference behaviour is materialised during the posterior inference stage and can be controlled by the injection of the associated inference type labels, $\Pi$, into the posterior. That is the conditional inference process:}
$$
P(x_{c}|x_{p_0}, x_{p_1}, \pi) = P(x_{c}|z^{(x_c)}) P(z^{(x_c)}|x_{p_0}, x_{p_1}, \pi)
$$
The inference type can be injected into the model at different points (e.g. at the encoder or decoder) and can be manipulated over different inference types to validate Proposition 1, as evaluated in Section \ref{sec:t5_inference_type}. Finally, optimising the language modelling task approximates the latent variable space $Z$. This can be formalised as: $P(x_c) = \prod_{i=1}^N P(c_i|c_{i-1}, \dots, c_1, Z)$
where $c_i$ represent the $i$-th token. The latent variable model can be used to deliver a symbolic-level control over the ``sentence'' space, i.e., syntactically independent spans of text \cite{li-etal-2022-variational-autoencoder}. Therefore, in the next section, we cover the sentence level representation $P(z|x_{p_0}, x_{p_1})$ and different architectural choices to encode the associated inference type controls. 

\subsection{Latent variable NLI architecture} \label{sec:latent_var_NLI_arc}
In this section, we describe the methodological framework behind the construction of the latent sentence space within T5, shown in Figure \ref{fig:architecture}. We first investigate the sentence representation setup, then, move to latent space injection.
\paragraph{Latent sentence space: $P(z|x_1, x_2)$.} While designing the sentence bottleneck, we compare the four most frequently used mechanisms to transform token embeddings into sentence embeddings: 

(1) Mean pooling: calculating the mean of each dimension on all token embeddings and feeding the resulting vector into a multi-layer perceptron to obtain the sentence embedding. (2) multi-layer perceptron (MLP): applying an MLP to reduce the dimensionality of token embeddings, and the resulting embeddings are concatenated to form a single sentence embedding: $z = \text{concat}\Big[ \text{MLP}_1(x_1); ...; \text{MLP}_T(x_T) \Big]$ where $\text{MLP}_i(x_i)$ represents the $i$-th neural network for input representation of token $x_i$, $z$ is the latent sentence representation, and $T$ is the maximum token length for a sentence. (3) multi-head attention: feeding each token embedding into the multi-head attention and considering the first output embedding as the sentence embedding \cite{montero2021sentence}: $z = \text{MultiHead}\left( XW^q, XW^k, XW^v \right)[0]$ where $X=[x_1, ..., x_T]$ and $W^q$, $W^k$, and $W^v$ are the weights for learning $q$, $k$, $v$ embeddings in self-attention, respectively. (4) Sentence T5: re-loading the pre-trained sentence T5 (S-T5, \citet{https://doi.org/10.48550/arxiv.2108.08877}).

\paragraph{Conditional generation:$P(x_c|z)$.} Next, we consider four strategies to inject sentence embeddings into the decoder. (1) Cross-attention input embedding (CA Input): reconstructing the token embeddings from a sentence representation and directly feeding them into the cross-attention layers of the decoder: $\hat{Y} = \text{MultiHead}\left(YW^q, \text{MLP}(z)W^k, \text{MLP}(z)W^v\right)$ where $\hat{Y}$ is the reconstruction of decoder input sequence $Y=[y_1, ..., y_K]$. (2) Cross-attention KV embedding (CA KV): instead of reconstructing the token embeddings, it consists of directly learning the Key and Value \cite{hu-etal-2022-fuse,li2020optimus}, which is formalised as $\hat{Y} = \text{MultiHead}\Big(YW^q, \text{MLP}_k(z), \text{MLP}_v(z)\Big)$, where $\text{MLP}_k$ and $\text{MLP}_v$ are neural layers for learning $k$ $v$ embeddings. (3) Non-cross-attention input connection (NCA Input): reconstructing the token embeddings and element-wisely adding them with the input embeddings of the decoder \cite{https://doi.org/10.48550/arxiv.2101.00828}. (4) Non-cross-attention output connection (NCA Output): adding the reconstructed token embeddings to the output embedding of the decoder.

\section{Empirical Analysis} \label{sec:empirical}


\subsection{Analysing the bottleneck baselines} \label{sec:t5_sentence_t5}
In this experiment, we analyze the performance of the sentence bottleneck choices on the explanatory NLI task (i.e., syllogistic-deductive NLI: generating one conclusion from two premises) and the explanatory inference retrieval task, using T5 as a reference model. All implementation details are provided in Appendix \ref{sec:hyper_param}.

\paragraph{Syllogistic-deductive NLI.} Firstly, we evaluate different encoding-decoding configurations on the performance of sentence bottleneck\footnote{We use the \textit{base} version as the baseline for T5 and sentence bottleneck T5.}.
As illustrated in Table \ref{tab:test_loss} (top), we can observe that the best configuration is that the encoder is sentence T5 (S-T5), and the decoder is connected by reconstructing token embeddings as the input of cross-attention. 

Next, we quantitatively evaluate the NLI performance of different baselines on the Entailment corpus. We specifically choose the VAE baselines, including the Transformer VAE model: Optimus \cite{li2020optimus} and Della \cite{hu-etal-2022-fuse} with two different sentence dimensions (32 and 768), and five LSTM language autoencoders with 768 latent dimensions: denoising AE (\citet{10.1145/1390156.1390294}, DAE), $\beta$-VAE \cite{Higgins2016betaVAELB}, adversarial AE (\citet{makhzani2016adversarial}, AAE), label adversarial AE (\citet{rubenstein2018latent}, LAAE), and denoising adversarial autoencoder (\citet{shen2020educating}, DAAE). In Table \ref{tab:test_loss} (bottom), we can observe that our T5bottleneck can outperform all baselines on BLEU \cite{Papineni02bleu:a}, BLEURT \cite{https://doi.org/10.48550/arxiv.2004.04696}, and cosine similarity from pre-trained sentence T5 \cite{https://doi.org/10.48550/arxiv.2108.08877}.

\begin{table}[ht!]
\centering
\resizebox{7.8cm}{!}{
\renewcommand\arraystretch{1}
\begin{tabular}{cccccc} \toprule
\multicolumn{6}{c}{\textit{Train: architecture}} \\
\multicolumn{2}{c}{Decoder Connection} & \shortstack{CA \\ Input} & \shortstack{CA \\ KV} & \shortstack{NCA \\ Input}  & \shortstack{NCA \\ Output}  \\ \hline
\multirow{4}{*}{\shortstack{Sentence \\ Embedding}}  &Pooling & 1.41 & 1.44 & 1.86  & 2.42  \\ 
&MLP & 1.71 & 1.94  & 2.09  & 2.62 \\ 
&MHA & 1.51 & 2.24 & 2.31  & 3.03 \\  
&S-T5 & \textcolor{blue}{\textbf{1.24}} & 1.42 & 1.81  & 2.22 \\ \hline \hline
\multicolumn{6}{c}{\textit{Test: EntailmentBank}} \\
Metrics & BLEU & Cosine & BLEURT & Loss $\downarrow$ & PPL $\downarrow$ \\ \hline
Optimus(32) & 0.07 & 0.74 & -1.20 & 1.13 & 2.31 \\
Optimus(768) & 0.08 & 0.74 & -1.21 & \textcolor{blue}{\textbf{0.82}} & \textcolor{blue}{\textbf{2.27}} \\
DELLA(32) & 0.08 & 0.85 & -1.23 & 1.69 & 5.41 \\
DELLA(768) & 0.09 & 0.87 & -1.09 & 1.54 & 4.66 \\ 
DAE(768) & 0.15 & 0.89 & -0.95 &  1.33 & 3.78\\
AAE(768) & 0.11 & 0.88 & -0.95 & 1.35 & 3.85 \\
LAAE(768) & 0.09 & 0.74 & -1.12 & 1.38 & 3.97\\ 
DAAE(768) & 0.07 & 0.74 & -1.20 & 1.43 & 4.17 \\ 
$\beta$-VAE(768) & 0.07 & 0.74 & -1.20 & 1.43 & 4.17\\ 
T5bottleneck & \textcolor{blue}{\textbf{0.35}} &\textcolor{blue}{\textbf{0.91}} & \textcolor{blue}{\textbf{-0.20}} & 1.24 & 3.45 \\ \toprule
\end{tabular}
}
\caption{Top: comparison of best setup on test loss via cross-entropy (CA: cross-attention, NCA: non-cross-attention), bottom: comparison of different baselines on EntailmentBank testset.} \label{tab:test_loss}
\end{table}


\begin{table}[ht!]
\resizebox{7.7cm}{!}{
\centering
\begin{tabular}{ccccc} \toprule
depth & t=1 & t=2 & t=3 & t=4 \\ \hline
DAE(768) & 30.27 & 31.74 & 30.65 & 30.74 \\
AAE(768) & 29.13 & 30.47 & 29.33 & 29.14 \\
LAAE(768) & 19.13 & 20.86 & 18.32 & 18.01 \\ 
DAAE(768) & 13.16 & 15.42 & 14.30 & 13.97 \\
$\beta$-VAE(768) & 10.03 & 10.07 & 10.05 & 10.05 \\
Optimus(768) & 28.21 & 29.35 & 28.35 & 28.27  \\ 
T5 bottleneck(768) & \textcolor{blue}{\textbf{34.47}} & \textcolor{blue}{\textbf{35.28}} & \textcolor{blue}{\textbf{34.50}} & \textcolor{blue}{\textbf{34.47}}  \\ \toprule
\end{tabular}
}
\caption{Explanatory inference retrieval task where t represents the depth of entailment tree.} \label{tab:scar}
\end{table}




\paragraph{Explanatory inference retrieval.} Furthermore, we evaluate the sentence embedding using as an associated explanation retrieval task (explanation-regeneration - i.e. retrieving the associated explanatory facts relevant to a claim) \cite{valentino2021hybrid}. Given a question-and-answer pair, it reconstructs the entailment tree by searching the explanations from a fact bank (i.e., WorldTree \cite{jansen-etal-2018-worldtree}) in an iterative fashion using a dense sentence encoder. In this framework, we can replace the dense sentence encoder with the proposed T5 bottleneck baseline to evaluate its sentence embeddings. We compare the T5 bottleneck with sentence VAEs: Optimus and five LSTM VAEs, and evaluate them via mean average precision (MAP). As illustrated in Table \ref{tab:scar}, the T5 bottleneck outperforms all baselines, indicating that it can deliver a better representation of explanatory sentences and entailment relations in a retrieval setting. 

\subsection{Analysing the impact of inference types} \label{sec:t5_inference_type}
Next, we evaluate the performance of the model with integrated inference types by comparing the following baselines: 1. seq2seq: T5 family and Bart \cite{https://doi.org/10.48550/arxiv.1910.13461}, 2. casualLM: GPT family \cite{Radford2019LanguageMA} and LlaMA3 and 3. seq2seq with sentence bottleneck: T5 bottleneck and Optimus. We focus on addressing two questions:

\noindent \textbf{1.} Can the inference types improve model training and inference? If so, models trained (or prompted) with inference types should exhibit higher scores (such as BLEU and BLEURT) on the test set.

\noindent \textbf{2.} Can inference types be used for a prescriptive inference control? If the association between inference types and the corresponding symbolic transformations are learned, manipulating the inference type could achieve a controlled generation. 

We consider three mechanisms to conditionally inject the inference types into the latent space, which are described below, where \textit{p1}, \textit{p2}, and \textit{con} are the premises and conclusion, respectively, and </s> is a special token for sentence separation.

\noindent \textbf{i.} The inference type as the prefix for the premises at the Encoder: \textit{the inference type is [type] </s> p1 </s> p2}

\noindent \textbf{ii.} The inference type as the prefix for the conclusion in the Decoder: \textit{</s> the inference type is [type]. con}

\noindent \textbf{iii.} The inference type at the end of the conclusion in the Decoder: \textit{</s> con. the inference type is [type]}

\paragraph{Fine-tuning.} We first quantitatively evaluate model performance based on five metrics: test loss (cross-entropy), perplexity (PPL), BLEURT, BLEU, and cosine similarity against sentenceT5. As illustrated in Table \ref{tab:metrics_with_inference_type}, all baselines with inference types always have lower test losses and PPLs, which means the inference type can help the model training. Furthermore, across all baseline models, incorporating inference types into the encoder consistently results in improved performance as measured by BLEU, Cosine, and BLEURT metrics. This finding suggests that the conditionalisation on inference types can support the inference representation, and the inference process has been performed inside the encoder (\textit{Proposition1}).

\begin{table}[ht!]
\resizebox{7.8cm}{!}{
\renewcommand\arraystretch{1}
\begin{tabular}{ccccccc} 
\toprule
Baseline & INJ & BLEU & Cosine & BLEURT & Loss $\downarrow$ & PPL $\downarrow$ \\ \hline
\multicolumn{7}{c}{\textit{seq2seqLM: encoder-decoder architecture}}  \\ \hline
\multirow{4}{*}{\shortstack{T5 \\ original \\ (small)}} & DE & 0.55 & 0.96 & 0.30 & 0.53 & 1.44 \\ 
& DP & 0.59 & 0.96 & 0.34 & 0.58 & 1.57 \\ 
& EP & \textcolor{blue}{\textbf{0.65}} & \textcolor{blue}{\textbf{0.97}} & \textcolor{blue}{\textbf{0.45}} & \textcolor{blue}{\textbf{0.52}} & \textcolor{blue}{\textbf{1.41}} \\
& NO & 0.54 & 0.96 & 0.22 & 0.69 & 2.22 \\ \hline

\multirow{4}{*}{\shortstack{T5 \\ original \\ (base)}} & DE & 0.46 & 0.96 & 0.23 & 0.49 & 1.33 \\ 
& DP & 0.53 & 0.96 & 0.25 & 0.51 & 1.38 \\ 
& EP & \textcolor{blue}{\textbf{0.61}} & \textcolor{blue}{\textbf{0.97}} & \textcolor{blue}{\textbf{0.39}} & \textcolor{blue}{\textbf{0.45}} & \textcolor{blue}{\textbf{1.22}} \\ 
& NO & 0.57 & 0.96 & 0.33 & 0.61 & 1.65 \\ \hline
\multirow{4}{*}{\shortstack{Bart \\ (base)}} & DE & 0.44 & 0.94 & 0.03 & 0.55 & 1.49 \\ 
& DP & 0.38 & 0.93 & -0.42 & \textcolor{blue}{\textbf{0.48}} & \textcolor{blue}{\textbf{1.30}} \\ 
& EP & \textcolor{blue}{\textbf{0.57}} & \textcolor{blue}{\textbf{0.96}} & \textcolor{blue}{\textbf{0.23}} & 0.58 & 1.57 \\ 
& NO & 0.54 & 0.96 & 0.17 & 0.63 & 1.71 \\ \hline 
\multirow{4}{*}{\shortstack{T5 \\ original \\ (large)}} & DE & 0.60 & 0.97 & 0.46 & \textcolor{blue}{\textbf{0.40}} & 1.49 \\ 
& DP & 0.64 & 0.97 & 0.44 & 0.46 & 1.58 \\ 
& EP & \textcolor{blue}{\textbf{0.67}} & \textcolor{blue}{\textbf{0.97}} & \textcolor{blue}{\textbf{0.50}} & 0.59 & 1.80 \\ 
& NO & 0.57 & 0.96 & 0.31 & 0.61 & 1.84 \\ \hline
\multirow{4}{*}{\shortstack{Flan-T5 \\ (large)}} & DE & 0.01 & 0.73 & -1.34 & 6.91 & 10.2 \\ 
& DP & 0.01 & 0.73 & -1.34 & 7.00 & 15.4 \\ 
& EP & \textcolor{blue}{\textbf{0.21}} & \textcolor{blue}{\textbf{0.87}} & \textcolor{blue}{\textbf{-1.04}} & \textcolor{blue}{\textbf{1.30}} & \textcolor{blue}{\textbf{3.66}} \\ 
& NO & 0.20 & 0.87 & -1.14 & 1.34 & 3.81 \\ \hline
\multirow{4}{*}{\shortstack{T5 \\ original \\ (3b)}} & DE & 0.60 & 0.96 & 0.44 & 0.68 & 1.97 \\ 
& DP & 0.66 & 0.96 & 0.49 & 0.65 & 1.91 \\ 
& EP & \textcolor{blue}{\textbf{0.70}} & \textcolor{blue}{\textbf{0.97}} & \textcolor{blue}{\textbf{0.57}} & \textcolor{blue}{\textbf{0.51}} & \textcolor{blue}{\textbf{1.66}} \\ 
& NO & 0.68 & 0.97 & 0.55 & 0.63 & 1.87 \\ \hline
\multicolumn{7}{c}{\textit{CausalLM: decoder only architecture}}  \\ \hline

\multirow{3}{*}{\shortstack{GPT2 \\ (large)}} & DE & 0.02 & 0.87 & -1.15 & 0.73 & 2.07 \\ 
& DP & \textcolor{blue}{\textbf{0.08}} & \textcolor{blue}{\textbf{0.90}} & \textcolor{blue}{\textbf{-0.91}} & \textcolor{blue}{\textbf{0.73}} & \textcolor{blue}{\textbf{2.07}} \\ 
& NO & 0.07 & 0.90 & -0.93 & 0.76 & 2.06 \\ \hline 

\multirow{3}{*}{\shortstack{GPT2 \\ (xl)}} & DE & 0.20 & 0.88 & -1.10 & 0.63 & 1.87 \\ 
& DP & \textcolor{blue}{\textbf{0.28}} & \textcolor{blue}{\textbf{0.91}} & \textcolor{blue}{\textbf{-0.90}} & \textcolor{blue}{\textbf{0.60}} & \textcolor{blue}{\textbf{1.82}} \\ 
& NO & 0.27 & 0.90 & -0.97 & 0.68 & 1.97 \\ \hline
\multicolumn{7}{c}{\textit{seq2seqLM with sentence bottleneck}}  \\ \hline
\multirow{4}{*}{\shortstack{T5 \\ bottleneck \\ (base)}} & DE & 0.35 & 0.91 & -0.15 & \textcolor{blue}{\textbf{0.84}} & \textcolor{blue}{\textbf{2.31}} \\ 
& DP & 0.39 & 0.91 & -0.13 & 0.86 & 2.36 \\ 
& EP & \textcolor{blue}{\textbf{0.42}} & \textcolor{blue}{\textbf{0.92}} & \textcolor{blue}{\textbf{-0.07}} & 1.23 & 3.42 \\
& NO & 0.35 & 0.91 & -0.20 & 1.24 & 3.45 \\ \hline
\multirow{4}{*}{\shortstack{Optimus \\ (BERT-GPT2)}} & DE & \textcolor{blue}{\textbf{0.26}} & \textcolor{blue}{\textbf{0.80}} & \textcolor{blue}{\textbf{-1.11}} & 0.87 & 2.38 \\ 
& DP & 0.25 & 0.79 & -1.14 & \textcolor{blue}{\textbf{0.85}} & \textcolor{blue}{\textbf{2.33}} \\ 
& EP & 0.09 & 0.74 & -1.17 & 1.11 & 3.03 \\
& NO & 0.07 & 0.74 & -1.20 & 1.13 & 3.09 \\
\toprule




\end{tabular}
}
\caption{Quantitative evaluation of test cases, where best results are highlighted in \textbf{\textcolor{blue}{bold}}. Specification for abbreviation. INJ: ways for injecting the information of inference types into the model, it includes DE: decoder end, DP: decoder prefix, EP: encoder prefix, NO: no inference type. PPL is perplexity, Loss is cross entropy.} \label{tab:metrics_with_inference_type}
\end{table}
\begin{table}[ht!]
\centering
\resizebox{7.8cm}{!}{
\begin{tabular}{cccccc}
\toprule
Baseline & INJ & Num & BLEU & Cosine & BLEURT \\ \hline
\multicolumn{6}{c}{\textit{Seq2seqLLM: encoder-decoder architecture}}  \\ \hline 
\multirow{4}{*}{\shortstack{CoT-T5 (11b) \\ \cite{kim2023cot}}}
& Yes & 10 & 0.51 & 0.97 & 0.39 \\
& Yes & 5 & 0.51 & 0.97 & 0.39 \\
& Yes & 0 & 0.50 & 0.97 & 0.36 \\ 
& NO & 0 & \textbf{\textcolor{red}{0.46}} & \textbf{\textcolor{red}{0.96}} & \textbf{\textcolor{red}{0.31}} \\ \hline
\multirow{4}{*}{\shortstack{Flan-T5 (xl)}}
& Yes & 10 & 0.49 & 0.96 & 0.40 \\
& Yes & 5 & 0.48 & 0.96 & 0.39 \\
& Yes & 0 & 0.52 & 0.96 & 0.39 \\ 
& NO & 0 & \textbf{\textcolor{red}{0.44}} & \textbf{\textcolor{red}{0.95}} & \textbf{\textcolor{red}{0.24}} \\ \hline
\multirow{4}{*}{\shortstack{Flan-T5 (xxl)}} 
& Yes & 10 & 0.51 & 0.97 & 0.41 \\
& Yes & 5 & 0.53 & 0.97 & 0.43\\
& Yes & 0 & 0.50 & 0.96 & 0.37 \\ 
& NO & 0 & \textbf{\textcolor{red}{0.48}} & \textbf{\textcolor{red}{0.96}} & \textbf{\textcolor{red}{0.36}} \\ \hline
\multicolumn{6}{c}{\textit{CausalLLM: decoder only architecture}}  \\ \hline
\multirow{4}{*}{\shortstack{GPT-3.5-turbo-0125}} 
& Yes & 10 & 0.52  & 0.96 & 0.40\\
& Yes & 5 & 0.48  & 0.96 & 0.35\\
& Yes & 0 &  0.46  & 0.96 &  \textbf{\textcolor{red}{0.31}}\\ 
& NO & 0 & \textbf{\textcolor{red}{0.42}} & 0.96 & 0.33 \\ \hline
\multirow{4}{*}{\shortstack{GPT-4-0613}} 
& Yes & 10 & 0.53 & 0.97 & 0.50 \\
& Yes & 5 & 0.52 & 0.97 & 0.47 \\
& Yes & 0 & 0.52 & 0.97 & 0.50 \\ 
& NO & 0 & \textbf{\textcolor{red}{0.47}} & \textbf{\textcolor{red}{0.96}} & \textbf{\textcolor{red}{0.40}} \\ \hline
\multirow{4}{*}{\shortstack{llama3-8b-8192}} 
& Yes & 10 & 0.48 & 0.96 & 0.33 \\
& Yes & 5 & 0.45 & 0.96 & 0.32 \\
& Yes & 0 & 0.37 & 0.95 & 0.22 \\ 
& NO & 0 & \textbf{\textcolor{red}{0.34}} & \textbf{\textcolor{red}{0.95}} & \textbf{\textcolor{red}{0.19}} \\ \hline
\multirow{4}{*}{\shortstack{llama3-70b-8192}} 
& Yes & 10 & 0.54 & 0.97 & 0.54 \\
& Yes & 5 & 0.52 & 0.97 & 0.52 \\
& Yes & 0 & 0.51 & 0.97 & 0.47 \\ 
& NO & 0 & \textbf{\textcolor{red}{0.44}} & \textbf{\textcolor{red}{0.96}} & \textbf{\textcolor{red}{0.40}} \\
\toprule
\end{tabular}
}
\caption{ICL evaluation of test cases, where worst results are highlighted in \textbf{\textcolor{red}{bold}}. The prompt is \textit{``performing natural language inference [where the inference type is type, description], $[p1; p2; c]_{\times \text{Num}}$. p1, p2, what is the conclusion?"}. The \textit{description} is based on the definition of inference types in Section \ref{sec:bg}.} \label{tab:metrics_llm}
\end{table}

\paragraph{In-context learning.} Next, we quantitatively evaluate the inference types within in-context learning (ICL) in contemporary large language models (LLMs). As illustrated in Table \ref{tab:metrics_llm}, prompting with inference types can improve the performance of ICL in both seq2seq and causal LLMs. Besides, within the context of causal LLMs, an increase in few shot examples\footnote{We randomly sample the examples with the same inference type as the current test example from the training set. We perform ten times and calculate the average for each metric.}, improves the performance. 

\paragraph{Controllability evaluation.} Moreover, we qualitatively\footnote{The quantitative evaluation cannot be automatically performed since some premises can only derive one conclusion.} evaluate the controllability on the generation of conclusions by systematically intervening on the inference type prior to the encoder. As illustrated in Table \ref{tab:control_generation}, we can observe that the associated linguistic properties of the conclusion can be controlled consistently with the inference type modifications, which indicates that the representation mechanisms can improve inference control with regard to symbolic/lexico-semantic properties. For example, when the type is ARG substitution (ARG-SUB), the model can generate \textit{the blacktop is made of a smooth surface} by replacing the argument \textit{asphalt concrete} with \textit{smooth surface}. The conclusions are changed to \textit{asphalt and blacktop have the same surface} when the inference type is the conjunction (FRAME-CONJ). More examples are provided in Table \ref{tab:more_example_3}.
\begin{table}[ht!]
\begin{tcolorbox}[fontupper=\small, fontlower=\small]
\underline{P1: \textcolor{blue}{blacktop} is made of \textcolor{red}{asphalt concrete}} \\
\underline{P2: \textcolor{red}{asphalt} has a \textcolor{orange}{smooth surface}}\\ \\
ARG-SUB: the \textcolor{blue}{blacktop} is made of \textcolor{orange}{smooth surface} \\
ARG-SUB-PROP: \textcolor{blue}{blacktop} has a \textcolor{orange}{smooth surface} \\
ARG/PRED-GEN: a \textcolor{blue}{blacktop} is a kind of \textcolor{red}{asphalt} \\
ARG-INS: \textcolor{red}{asphalt concrete} \textcolor{blue}{blacktop} has a \textcolor{orange}{smooth surface} \\
FRAME-CON: \textcolor{red}{asphalt} and \textcolor{blue}{blacktop} have the same surface \\
IFT: if the \textcolor{red}{asphalt} has a \textcolor{orange}{smooth surface} then the \textcolor{blue}{blacktop} will have a \textcolor{orange}{smooth surface}
\end{tcolorbox}

\caption{Controllable generation over original T5 (base) (ARG-SUB: argument substitution, ARG/PRED-GEN: argument/predicate generalisation. ARG-SUB-PROP: property inheritance. ARG-INS: argument insertion, FRAME-CON: frame conjunction, IFT: if...then...). The example of the T5 bottleneck is provided in Table \ref{tab:control_generation_comparison}.}
\label{tab:control_generation}
\end{table}

\section{Conclusion and Future Work} \label{sec:concl}
In this work, we focus on the syllogistic-deductive NLI inference task to explore symbolic-level inference control on language models. We first define a set of universal inference types for capturing inference rules behind explanatory sentences, grounded on AMR graphs, as a mechanism to precisely define symbolic inference operations. 

To integrate symbolic-level annotations into the NLI model, we formalise the neural NLI model as a latent variable model, enabling a bridge between symbolic and distributional representations. Based on this framework, our symbolic inference control is modelled as learning a conditional latent variable model. Besides, as a complementary, we propose T5 Bottleneck architecture to enhance the research on sentence representations.ich 

Experimental results revealed that the proposed approach can improve model training, inference, in-context learning, retrieval, and symbolic control. Based on our findings, in the future, we will explore the dynamics of symbolic explanatory inference patterns inside language models to enhance their mechanism interpretability in the NLI task \cite{deng2021discovering,li2023does}. 


\section{Limitations}

While the work focuses on the symbolic control of explanatory inference, complementary methods need to be employed to deliver more strict safety guarantees. In short, control and safety should not be confounded and should be delivered by independent mechanisms.

Quantitative evaluation of controllability depends on the elaboration of a new ground-truth dataset, mapping the valid conclusions to a set of premises and the corresponding inference types. This is a direction to be explored in future work.

\bibliography{references}
\bibliographystyle{acl_natbib}

\appendix
\clearpage
\appendix
\section{Annotation details} \label{sec:annotation}
\paragraph{Other inference types.} the \textit{other inference types} category includes (9) \textit{example}, (10) \textit{if-then}, and (11) \textit{others}. Those inference types are defined according to the key lexical characteristic of the conclusion, as systematic AMR transformations which could be applied without rephrasing the underlying explanatory sentences could not be determined. For example, the \textit{if-then} category refers to a conclusion that has an \textit{if then} sentence structure in which both premises do not contain it. The \textit{example} category is a conclusion with the signalling word \textit{example}, which does not appear in either premise. As for other unknown categories, we do not further specify them, as they either require knowledge outside of the scope of the premises or do not have a consistent symbolic transformation expression. An additional subtype called \textit{premise copy} was included for the cases where the conclusion has the same graph as one of the premises. 

\paragraph{Annotation procedure.} Annotation was performed manually for 5134 entailment triples (two premises, one conclusion) from the EntailmentBank \cite{dalvi2021explaining}, according to Algorithm \ref{alg:annotation}. Graph subset relations and root matching are relaxed for non-argument (:ARG*, op*) edges, meaning relations such as \textit{:manner} or \textit{:time} can be ignored for this purpose. Two independent annotators with post-graduate level backgrounds in Computational Linguistics were used in this process, on a consensus-based annotation scheme where a first annotator defined the transformations and a second annotator verified and refined the annotation scheme, in two iterations. The annotation of the AMR graph is based on an off-the-shelf parser \cite{damonte-17}. The descriptions for each inference type category are as follows:

\textbf{ARG-SUB} (Figure \ref{fig:amr_argsub}): the conclusion is obtained by replacing one argument with another argument.

\textbf{PRED-SUB}: the conclusion is obtained by replacing one verb with another verb.

\textbf{FRAME-SUB}: the conclusion is obtained by replacing a frame of one of the premises with one from the other premise.

\begin{figure}[ht!]
    \centering
    \includegraphics[width=\columnwidth]{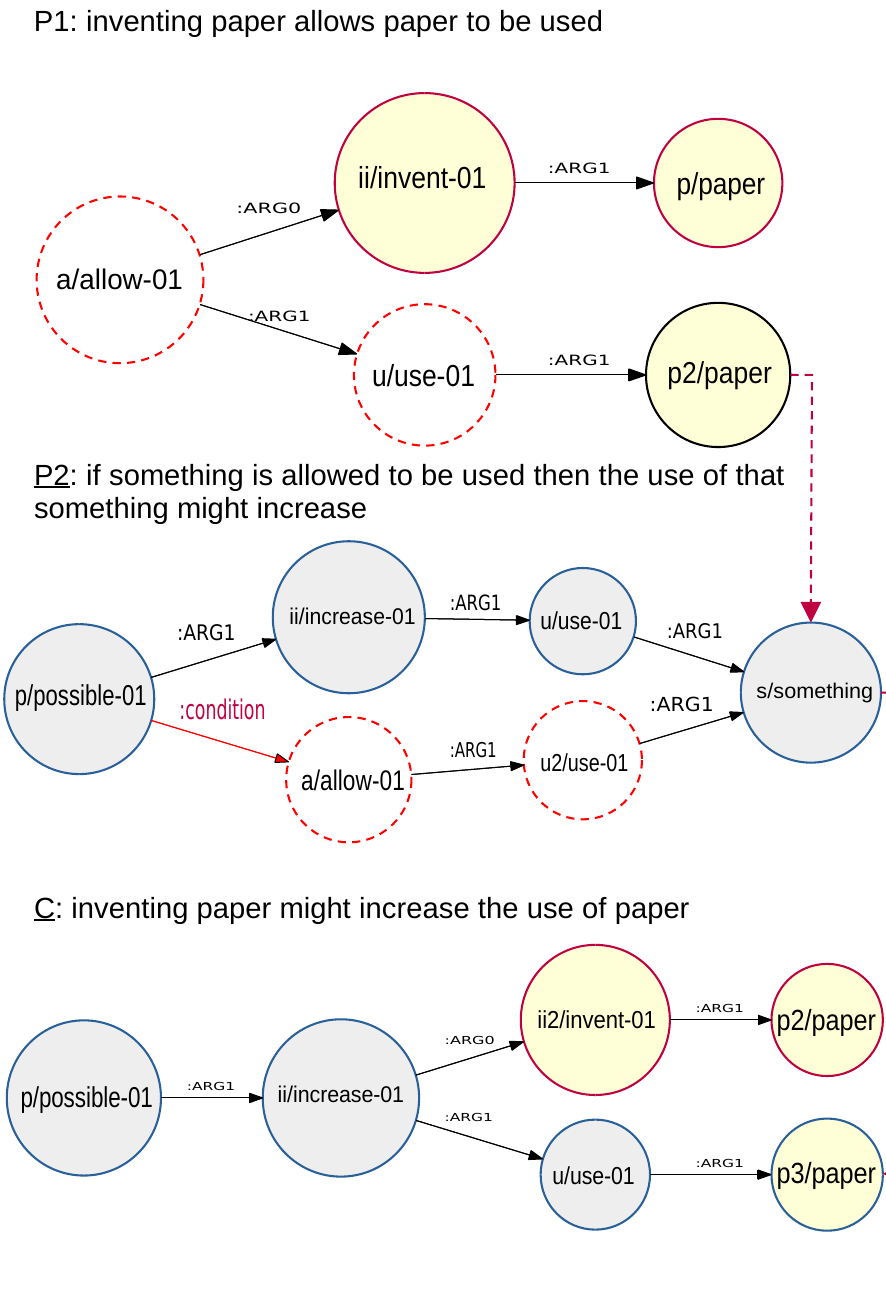}
    \caption{AMR conditional frame insertion (COND-FRAME).}
    \label{fig:amr_condframesub}
\end{figure}

\textbf{COND-FRAM} (Figure \ref{fig:amr_condframesub}): the conclusion is obtained according to the conditional premise with keyword ``if".

\begin{figure}[ht!]
    \centering
    \includegraphics[width=\columnwidth]{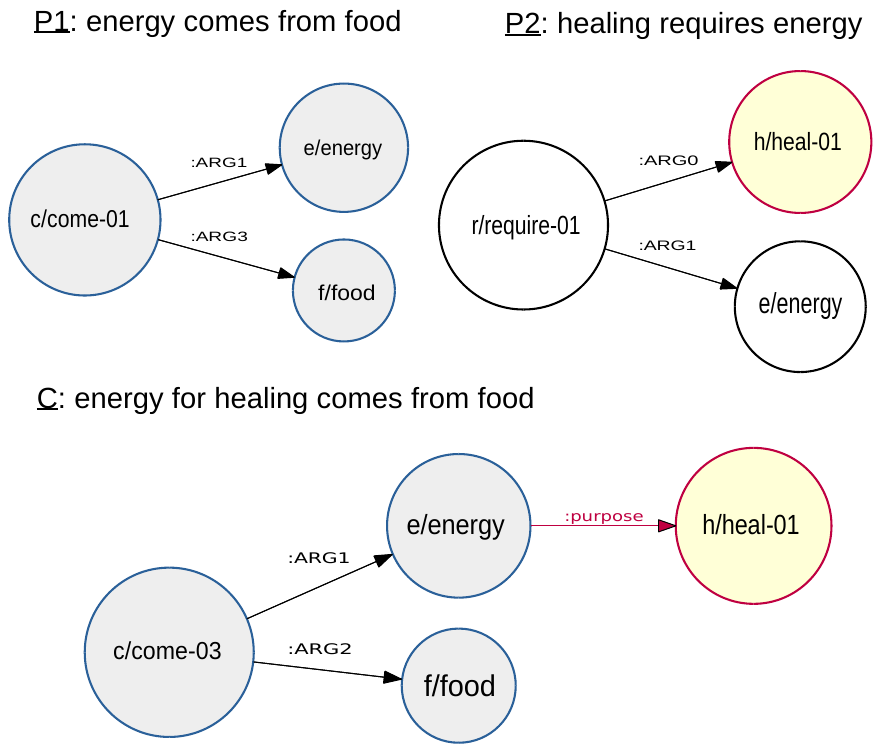}
    \caption{AMR argument insertion (ARG-INS).}
    \label{fig:amr_arginsert}
\end{figure}

\textbf{ARG-INS} (Figure \ref{fig:amr_arginsert}): the conclusion is obtained by connecting an argument from one of the premises to a frame of the other.

\textbf{FRAME-CONJ}: the conclusion is obtained by using connectives to connect two premises.

\begin{figure}[ht!]
    \centering
    \includegraphics[width=\columnwidth]{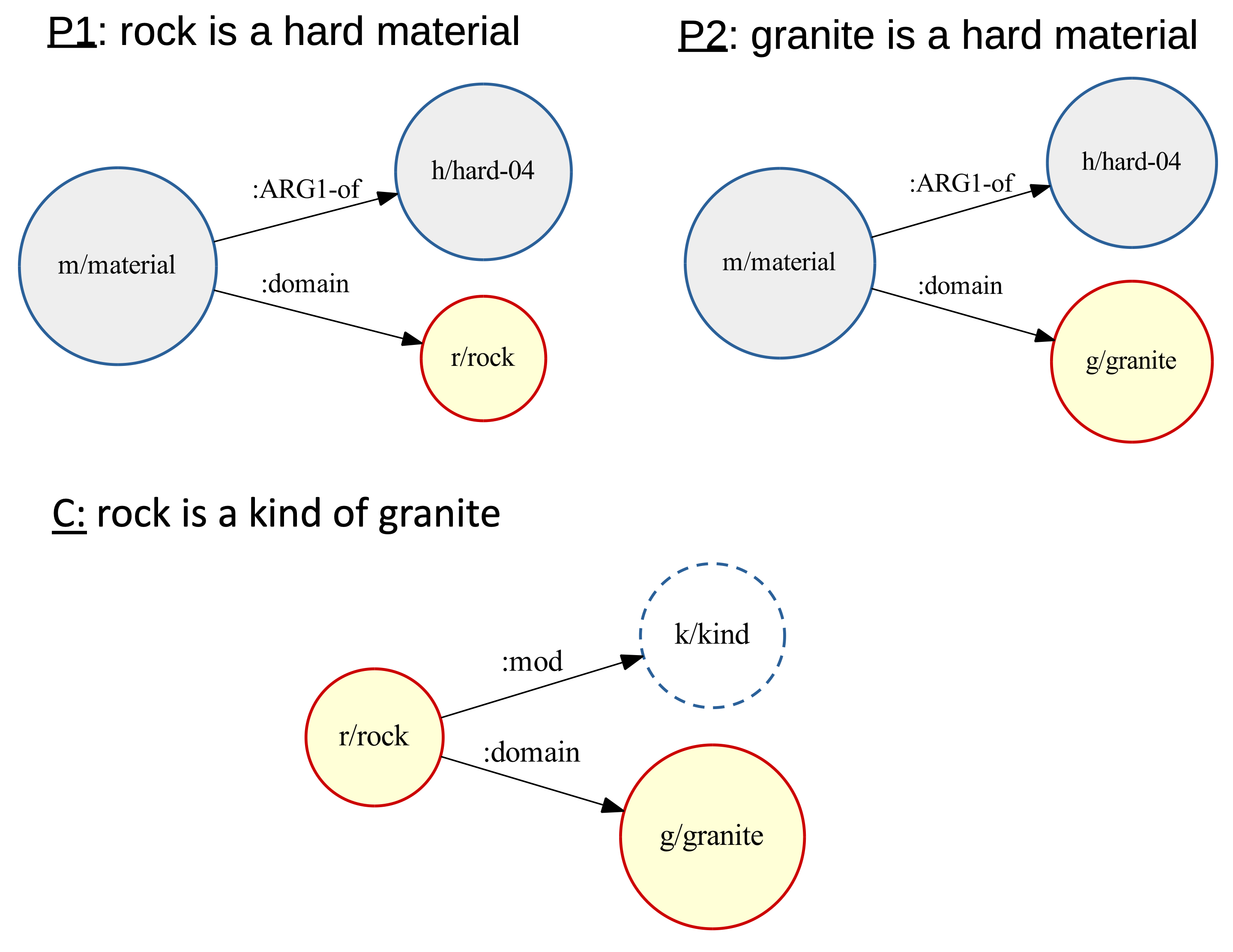}
    \caption{AMR argument generalisation (ARG-GEN).}
    \label{fig:amr_argpredgeneralis}
\end{figure}

\textbf{ARG/PRED-GEN} (Figure \ref{fig:amr_argpredgeneralis}): a new \textit{:domain} relation frame is created in the conclusion if both premise graphs differ by a single predicate/argument term.

\begin{figure}[ht!]
    \centering
    \includegraphics[width=\columnwidth]{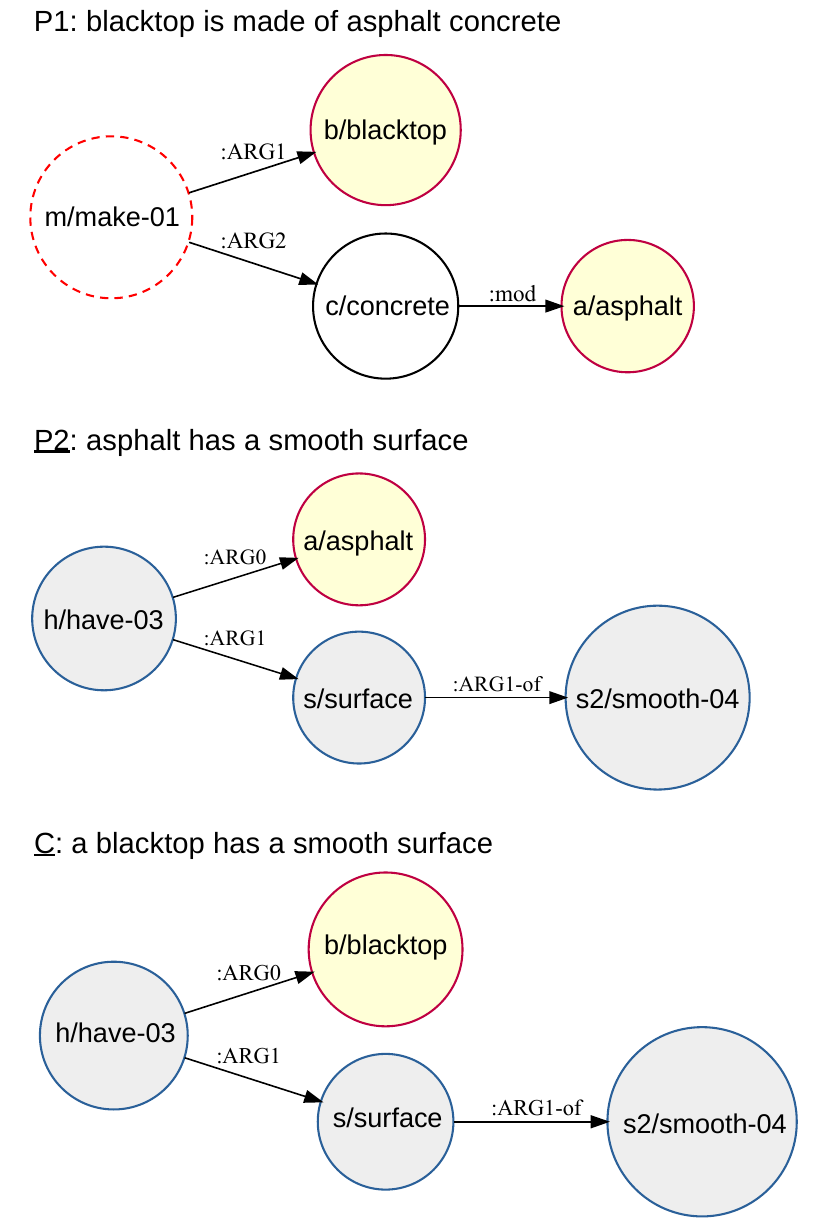}
    \caption{AMR argument substitution (property inheritance) (ARG-SUB-PROP).}
    \label{fig:amr_argsubprop}
\end{figure}

\textbf{ARG-SUB-PROP} (Figure \ref{fig:amr_argsubprop}): one of the premises describes a ``\textit{is made of}'' relationship between the entity in the other premise and its replacement.

\textbf{IFT}: the conclusion should be a conditional sentence.

\textbf{EXAMPLE}: the conclusion should contain the keyword ``example".

\section{Implementation details} \label{sec:hyper_param}
\begin{figure*}[ht!]
\centering
\includegraphics[width=15cm]{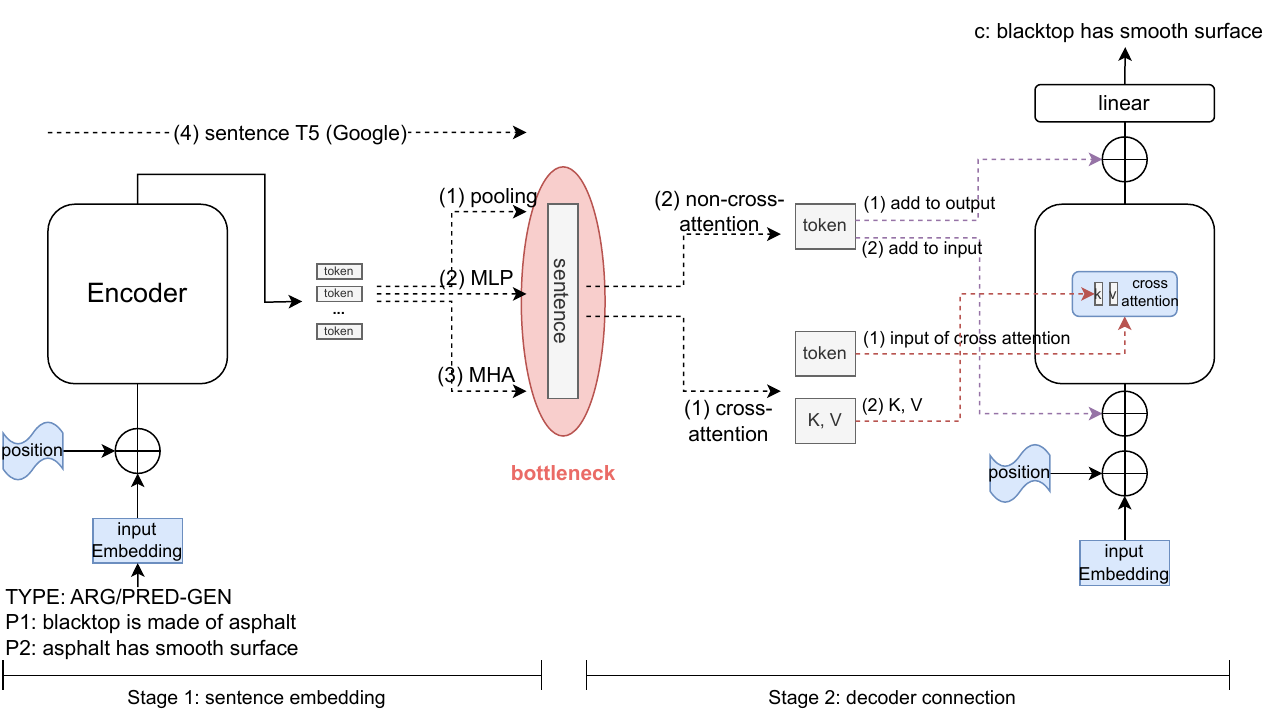}
\caption{The architectural configuration of T5bottleneck.}
\label{fig:architecture}
\end{figure*}

\paragraph{Architecture overview.} Figure \ref{fig:architecture} shows the architecture of T5 bottleneck for learning latent sentence space. It includes two stages: sentence embedding and decoder connection. The sentence embedding aims to transform token embeddings into a sentence (single) embedding. Decoder connection aims to connect the encoder and decoder.

\paragraph{Implementation details.} \textbf{1.} Size of Sentence Representation: in this work, we consider 768 as the size of the sentence embedding. Usually, the performance of the model improves as the size increases. \textbf{2.} Multi-head Attention (MHA): in the experiment, MHA consists of 8 layers, each layer containing 12 heads. The dimensions of Query, Key, and Value are 64 in each head. The dimension of token embedding is 768. Training hyperparameters are: \textbf{3.} Max epoch: 40, learning rate: 5e-5. During fine-tuning the sentence bottleneck T5, we first freeze the pre-trained parameters in the first epoch and fine-tune all parameters for the remaining epochs. \textbf{4.} The implementation of Optimus is based on their original code \cite{li2020optimus} \footnote{\url{https://github.com/ChunyuanLI/Optimus}}. It is initialised with a pretrained checkpoint with $\beta$=0.
\begin{table}[ht!]
\centering
\scriptsize
\resizebox{7.8cm}{!}{
\centering
\begin{tabular}{ccccccc}
\toprule
Size & 32 & 64 & 128 & 256  & 512 & 768 \\ \hline
Loss & 1.59 & 1.42 & 1.36 & 1.32 & 1.29 & 1.24 \\ \toprule
\end{tabular}
}

\caption{The loss with different sentence dim in sentence bottleneck T5.} \label{tab:arg0_exp}
\end{table}

\paragraph{Baselines.} In the experiment, we implement five LSTM-based autoencoders, including denoising AE (\citet{10.1145/1390156.1390294}, DAE), $\beta$-VAE \cite{Higgins2016betaVAELB}, adversarial AE (\citet{makhzani2016adversarial}, AAE), label adversarial AE (\citet{rubenstein2018latent}, LAAE), and denoising adversarial autoencoder (\citet{shen2020educating}, DAAE). Their implementation relies on the open-source codebase available at the URL \footnote{\url{https://github.com/shentianxiao/text-autoencoders}}. As for transformer-based VAEs, we implement Optimus \cite{li2020optimus} and Della \cite{hu-etal-2022-fuse}\footnote{\url{https://github.com/OpenVLG/DELLA}}. All baseline models undergo training and evaluation with the hyper-parameters provided by their respective sources. A latent dimension of 768 is specified to ensure a uniform and equitable comparative analysis.

\paragraph{Metrics.} To compare the generation with the golden conclusion, we choose the BLEU with 1-3 grams and report the average score. As for the cosine similarity from sentence T5, we feed both conclusions into pretrained sentence T5 \footnote{\url{https://huggingface.co/sentence-transformers/sentence-t5-base}} and report the cosine similarity.  

\section{Complementary Results} \label{sec:ablation}
\paragraph{Ablation studies.} We remove the inference types from the dataset and evaluate the T5 model performance using the same metrics. In this case, we can compare the model performance trained with or without that inference-type. From Table \ref{tab:ablation_study}, we can observe that the baselines (T5 small and base) achieve higher BLEU and BLEURT scores without the data with ARG-INS, COND-FRAME, and UNK inference type, respectively. This result indicates that the T5 cannot generalize well over those inference types. Also, removing the UNK inference type from data can achieve lower loss and PPL, which indicates that it has a negative impact on model training. 
\begin{table}[ht!]
\centering
\setlength\tabcolsep{2.5pt}
\resizebox{7.8cm}{!}{
\centering
\begin{tabular}{ccccccc} \toprule
Remove & T5 & BLEU & BLEURT & Cosine & Loss $\downarrow$ & PPL $\downarrow$ \\ \hline
\multirow{2}{*}{\shortstack{FRAME-\\SUB}} & small & 0.50 & 0.19 & 0.95 & 0.95 & 2.58 \\ 
& base & 0.60 & 0.33 & 0.96 & 0.72 & 1.95 \\ \hline

\multirow{2}{*}{ARG-INS} & small & \textcolor{blue}{\textbf{0.54}} & \textcolor{blue}{\textbf{0.27}} & 0.95 & 0.82 & 2.22 \\ 
& base & \textcolor{blue}{\textbf{0.63}} & \textcolor{blue}{\textbf{0.46}} & 0.97 & 0.64 & 1.73 \\ \hline

\multirow{2}{*}{\shortstack{FRAME-\\CONJ}} & small & 0.53 & 0.26 & 0.96 & 0.84 & 2.28 \\ 
& base & 0.60 & 0.35 & 0.96 & 0.65 & 1.76 \\ \hline

\multirow{2}{*}{\shortstack{COND-\\FRAME}} & small & \textcolor{blue}{\textbf{0.55}} & \textcolor{blue}{\textbf{0.25}} & 0.96 & 0.88 & 2.39 \\ 
& base & \textcolor{blue}{\textbf{0.59}} & \textcolor{blue}{\textbf{0.36}} & 0.96 & 0.69 & 1.87 \\ \hline

\multirow{2}{*}{UNK} & small & \textcolor{blue}{\textbf{0.55}} & \textcolor{blue}{\textbf{0.23}} & 0.95 & \underline{0.53} & \underline{1.44} \\ 
& base & \textcolor{blue}{\textbf{0.62}} & \textcolor{blue}{\textbf{0.40}} & 0.96 & \underline{0.58} & \underline{1.57} \\ \hline

No & small & 0.54 & 0.22 & 0.96 & 0.69 & 2.22 \\ 
No & base & 0.57 & 0.33 & 0.96 & 0.61 & 1.65 \\ \toprule

\end{tabular}
}
\caption{Ablation study over inference type (No: no inference types are removed).} \label{tab:ablation_study}
\end{table}

\paragraph{More controllable inference examples.} \label{sec:example_control}
We provide more controlled examples based on both the Original T5 and T5 bottleneck in Table \ref{tab:control_generation_comparison}, \ref{tab:control_generation_comparison1}, and \ref{tab:more_example_3}. All examples reveal that the inference type can provide quasi-symbolic inference control to language models.
\begin{table}[ht!]
\begin{tcolorbox}[fontupper=\small, fontlower=\small]
\underline{P1: a \textcolor{orange}{pumpkin} contains \textcolor{blue}{seeds}} \\
\underline{P2: \textcolor{green}{fruit} contains \textcolor{blue}{seeds}}\\

Original T5: \\
ARG-INS: a \textcolor{green}{fruit} in a \textcolor{orange}{pumpkin} contains \textcolor{blue}{seeds} \\
FRAME-CONJ: a \textcolor{orange}{pumpkin} and \textcolor{green}{fruit} both contains \textcolor{blue}{seeds} \\
FRAME-SUB: \textcolor{green}{fruit} is a kind of \textcolor{orange}{pumpkin}
\tcblower
T5 bottleneck: \\
ARG-INS: \textcolor{green}{fruit} is a part of \textcolor{orange}{pumpkin} that contains \textcolor{blue}{seeds} \\
FRAME-CONJ: a \textcolor{green}{fruit} contains \textcolor{blue}{seeds} \\
FRAME-SUB: a \textcolor{orange}{pumpkin} is a kind of plant
\end{tcolorbox}
\caption{Controlled generation. original T5(base) (top) and T5 bottleneck (bottom).}
\label{tab:control_generation_comparison}
\end{table}

\begin{table*}[ht!]
\begin{tcolorbox}[fontupper=\small, fontlower=\small]
    \underline{P1: eating \textcolor{blue}{something} has a negative impact on} \underline{\textcolor{blue}{that something}} \\
    \underline{P2: some \textcolor{red}{animals} eat \textcolor{green}{cacti}} \\
    ARG-INS: some \textcolor{red}{animals} have a negative impact on \textcolor{green}{cacti} by eating \textcolor{green}{cacti} \\
    PRED-SUB: some \textcolor{red}{animals} may have a negative impact on
    \textcolor{green}{cacti} \\
    FRAME-SUB: eating \textcolor{green}{cacti} has a negative impact on that \textcolor{green}{cacti}
    \tcblower
    ARG-INS: some \textcolor{red}{animals} have a negative impact on \textcolor{green}{cacti} by eating \textcolor{green}{cacti} \\
    PRED-SUB: \textcolor{red}{animals} have a negative impact on \textcolor{green}{cacti} \\
    FRAME-SUB: eating \textcolor{green}{cacti} has a negative impact on that \textcolor{green}{cacti}
\end{tcolorbox}
\caption{Controlled generation. original T5(base) (top) and T5 bottleneck (bottom).}
\label{tab:control_generation_comparison1}
\end{table*}

\begin{table*}[ht!]
\scriptsize
\centering
\resizebox{15.6cm}{!}{
\begin{tabular}{p{4.2cm}p{3.2cm}p{8.2cm}}
\toprule
Premises & Inference Type & Original T5  \\ \hline
P1: a pumpkin contains seeds & ARG-INS & a fruit in a pumpkin contains seeds  \\ 
P2: fruit contains seeds & FRAME-CONJ & a pumpkin and fruit both contain seeds \\ 
 & IFT & if a pumpkin contains fruit then the fruit may contain seeds \\ 
 & EXAMPLE & fruit is an example of pumpkins being sown \\
 & ARG/PRED-GEN & a pumpkin is a kind of fruit \\
 & ARG-SUB & fruit can contain pumpkin seeds \\
 & UNK & a pumpkin can contain seeds \\
 & FRAME-SUB & fruit is a kind of pumpkin \\ \hline
P1: sunlight is a kind of solar energy & ARG-INS & solar energy is a kind of resource for plants that uses water \\
P2: water and sunlight are resources for plants & FRAME-CONJ & water and sunlight are resources for plants and are kinds of solar energy \\ 
& UNK & the resources for plants include water and solar energy \\
& ARG-SUB & water and solar energy are resources for plants \\ \hline
P1: to move something can mean to transfer something & ARG-SUB & flowing can mean to transfer energy \\
P2: flowing is a kind of movement for energy & INF & if something flows, then that energy will flow \\
& FRAME-CONJ & moving can transfer energy and mean flowing \\
& ARG-INS & flowing can be a kind of transfer of energy to another entity \\ 
& ARG/PRED-GEN & transferring energy with flowing can be seen as transferring energy \\
P1: if a pot is exposed to a stove then that pot may become hot & COND-FRAME & the pot will become hot \\ 
P2: the pot is exposed to a stove & EXAMPLE & an example of hot pot is a pot exposed to a stove \\ \hline
P1: eating something has a negative impact on that something & FRAME-SUB & eating cacti has a negative impact on that cacti \\
P2: some animals eat cacti & PRED-SUB & some animals may have a negative impact on cacti \\
& ARG-INS & some animals have a negative impact on cacti by eating cacti \\
& EXAMPLE & cooking cacti is an example of a negative impact on a cactus \\
& INF & if a cactus has a negative impact on an animal, that cactus could be devoured \\ \hline
P1: seeing requires light & ARG-SUB & reading requires light \\
P2: reading requires seeing & ARG-INS & light is a kind of requirement for reading \\
& INF & if light is moving then reading may be taken \\
& EXAMPLE & a light bulb will be used for reading \\
& UNK & light will help you read \\
\toprule
\end{tabular}
}
\resizebox{15.6cm}{!}{
\begin{tabular}{p{6.2cm}p{3.2cm}p{6.2cm}}
\toprule
Premises & Inference Type & T5 bottleneck  \\ \hline
P1: a pumpkin contains seeds & ARG-INS & fruit is part of a pumpkin that contains seeds  \\ 
P2: fruit contains seeds & FRAME-CONJ & a fruit contains seeds \\ 
 & FRAME-SUB & a pumpkin is a kind of plant \\ \hline
P1: sunlight is a kind of solar energy & ARG-INS & water is a kind of resource that is used by plants for growth \\
P2: water and sunlight are resources for plants & FRAME-CONJ & plants and water are resources that require water and energy \\ 
& ARG-SUB & plants use water and sunlight to produce energy \\ \hline
P1: to move something can mean to transfer something & ARG-SUB & flowing can mean to transfer energy \\
P2: flowing is a kind of movement for energy & INF & if something flows, then that energy will flow \\
& FRAME-CONJ & moving can transfer energy and mean flowing \\
& ARG-INS & flowing can be a kind of transfer of something \\ 
& ARG/PRED-GEN & transferring energy with flowing can be seen as transferring energy \\ \hline
P1: if a pot is exposed to a stove then that pot may become hot & COND-FRAME & the pot may become hot \\ 
P2: the pot is exposed to a stove & ARG/PRED-GEN & the pot may be a source of heat \\ \hline
P1: eating something has a negative impact on that something & FRAME-SUB & eating cacti has a negative impact on that cacti \\
P2: some animals eat cacti & PRED-SUB & animals have a negative impact on cacti \\
& ARG-INS & some animals have a negative impact on cacti by eating cacti \\ \hline

P1: seeing requires light & ARG-SUB & reading requires light \\
P2: reading requires seeing & FRAME-CONJ & reading and feeling can both be used \\
& INF & if something is visible then that something will be seen \\ \toprule
\end{tabular}
}
\caption{controllable NLI via inference type (Top: original T5, bottom: T5 bottleneck).} \label{tab:more_example_3}
\end{table*}

\begin{algorithm*}[ht!]
    \caption{Annotation procedure} \label{alg:annotation}
    \begin{algorithmic}[1]
    \State Find premise $P_x$ most similar to the conclusion $C$, $P_{\bar{x}}$ being the other premise.
    \State $G_{x, \bar{x}, C}~~\gets~~ $AMR~graph~of~$P_x, P_{\bar{x}}, C$,~~respectively.
    \State \textcolor{blue}{\# - - - - - - - - - - - - - - - - - - - common ARG-SUB, PRED-SUB - - - - - - - - - - - - - - - - - - - - - - - - - -}
    \If{$G_x = G_c$ or $G_{\bar{x}} = G_c$}
        \State $type = PREM\textnormal{-}COPY$ \textcolor{gray}{\# Comment: no reasoning happen.}
    \ElsIf{$P_x$ and $C$ differ by one word $w$} \textcolor{gray}{\# Comment: common ARG(PRED)-SUB.}
        \If{$w$ is a verb}
            \State $type = PRED\textnormal{-}SUB$
        \Else
            \State $type = ARG\textnormal{-}SUB$
        \EndIf
    \Else \\
    \textcolor{blue}{\# - - - - - - - - - - - - - - - - - COND-FRAME, FRAME-SUB, ARG-SUB-PROP - - - - - - - - - - - - - - }
        \State Get AMR graphs $G_1$, $G_2$, $G_c$ for $P_1$, $P_2$ and $C$ respectively. $P_x \rightarrow G_x$.
        \If{$\exists~\textnormal{:ARG*}(x, a)$ $\in C$ and $a \in P_{\bar{x}}$}
            \If{$\exists$ :condition($root(G_x)$, $root(G_{\bar{x}})$)} \\
            ~~~~~~~~~~ \textcolor{gray}{\# Comment: see Figure \ref{fig:amr_condframesub}, two root nodes are connected by :condition edge}
                \State $type = COND\textnormal{-}FRAME$ 
            \ElsIf{$root(a)$ is a noun}
                \If{$root(G_{\bar{x}}) =$ ``make-01'' and $\exists$ :ARG*($root(G_{\bar{x}})$, a)} \\
                ~~~~~~~~~~ \textcolor{gray}{\# Comment: ``make'' as a trigger to classify ARG-SUB and property inheritance.}
                    \State $type = ARG\textnormal{-}SUB\textnormal{-}PROP$
                \Else
                    \State $type = ARG\textnormal{-}SUB$ \textcolor{gray}{\# ARG-SUB that was not caught by the simpler rule on line 10, due to Px differing from C by more than a single word}
                \EndIf
            \Else
                \State $type = FRAME\textnormal{-}SUB$
            \EndIf \\
        \textcolor{blue}{\# - - - - - - - - - - - - - -  - - - - - Further-specification and Conjunction - - - - - - - - - - - - - - - - - - - - - - -}
        \ElsIf{$G_x \subset G_c$ and $G_{\bar{x}} \subset G_C$}
            \State $type = FRAME\textnormal{-}CONJ$
        \ElsIf{$\exists x, y$ :domain($root(G_x)$, $x$) and :domain($root(G_{\bar{x}}$, $y$) and :op*(``and'', $x$) $\in G_c$ and :op*(``and'', y) $\in G_c$} \textcolor{gray}{\# Comment: using connectives `and' to connect two premises}
            \State $type = FRAME\textnormal{-}CONJ$
        \ElsIf{$G_x \subset G_c$}
            \State $d \gets G_c - G_x$
            \If{$root(d)$ is a noun}
                \State $type = ARG\textnormal{-}INS$ \textcolor{gray}{\# Comment: inserting an argument.}
            \Else
                \State $type = FRAME\textnormal{-}INS$ \textcolor{gray}{\# Comment: inserting a phase (also annotated as ARG-INS).}
            \EndIf \\
        \textcolor{blue}{\# - - - - - - - - - - - - - - - - - - - - ARG/PRED-GEN and Others - - - - - - - - - - - - - - - - - - - - - - -  - - - - }
        \ElsIf{$\exists$ :domain($root(G_c)$, $y$) and ($root(G_c) \in G_x$ and $y \in G_{\bar{x}}$) or ($root(G_c) \in G_{\bar{x}}$ and $y \in G_x$)}
            \State $type = ARG/PRED\textnormal{-}GEN$
        \Else
            \State $type = UNK$
        \EndIf
    \EndIf
    \end{algorithmic}
\end{algorithm*}

\end{document}